\documentclass[letterpaper, 10pt, journal, twoside]{IEEEtran}
\usepackage{amsmath,amsfonts}
\usepackage{array}
\usepackage[caption=false,font=normalsize,labelfont=sf,textfont=sf]{subfig}
\usepackage{textcomp}
\usepackage{stfloats}
\usepackage{url}
\usepackage{verbatim}
\usepackage{graphicx}

\def\BibTeX{{\rm B\kern-.05em{\sc i\kern-.025em b}\kern-.08em
    T\kern-.1667em\lower.7ex\hbox{E}\kern-.125emX}}
\usepackage{balance}
\usepackage{algorithm}
\usepackage{algpseudocode}
\usepackage{booktabs}
\usepackage[colorlinks=true, linkcolor={rgb:red,2;green,1;blue,0.5}, citecolor=blue, urlcolor=red]{hyperref}
\usepackage{multirow}
\usepackage{tabularx}
\usepackage{xcolor} 
\usepackage[T1]{fontenc} 
\usepackage{xcolor}
\begin{document}
\title{Robust Fall Recovery for Armless Bipedal-Wheeled Robots via Force‑Guided Learning}
\author{Haidong Hou, Zhangguo Yu, Tao Han, Hengbo Qi, Khaleel Ghazal, 
Yu Zhang, \\ Yidong Du, Xuechao Chen, and Fei Meng%
\thanks{Manuscript received: January 28, 2026; Revised: March 24, 2026; Accepted: May 14, 2026.}%
\thanks{This paper was recommended for publication by Editor Lucia Pallottino upon evaluation of the Associate Editor and Reviewers comments.}%
\thanks{This work was supported in part by the National Natural Science Foundation of China under Grant 52575004, and in part by the Beijing Natural Science Foundation under Grants L252015 and L243004.}
\thanks{The authors are with the School of Mechatronical Engineering, Beijing Institute of Technology, Beijing 100081, China (e-mail: houhaidong@bit.edu.cn; corresponding author: yuzg@bit.edu.cn).}%
\thanks{Digital Object Identifier (DOI): see top of this page.}
}

\markboth{IEEE ROBOTICS AND AUTOMATION LETTERS. PREPRINT VERSION. ACCEPTED MAY, 2026}%
{Hou \MakeLowercase{\textit{et al.}}: Robust Fall Recovery for Armless Bipedal-Wheeled Robots via Force‑Guided Learning}

\maketitle

\begin{abstract} 
Fall recovery is critical for autonomous legged locomotion. Existing methods have demonstrated that some legged robots, such as humanoids and quadrupeds, are capable of fall recovery from diverse postures by utilizing arms or coordinating multi-legs to generate support forces. Without arms or other legs to provide supportive assistance, a bipedal-wheeled robot must rely solely on the actuation of its legs, making recovery particularly difficult. To address this, we introduce FTSR (Force-guided Teacher-student framework with Stage-wise Rewards). The force-guided method constructs an external auxiliary force during simulation training that correlates directly with the robot's real-time height, explicitly formulating this force as an optimizable constraint. Through constrained reinforcement learning, the policy is guided toward reducing force dependency gradually and increasing the body height, developing internal recovery strategies despite having no arms for support. Height-progressive stage-Wise rewards progressively structure posture stabilization during recovery and transition to sustained locomotion, integrated with teacher-student architecture distilling privileged knowledge of force effects and recovery dynamics. After simulation training, the policy is deployed on a physical armless bipedal-wheeled robot and extensively evaluated. Experiments confirm robust and reliable fall recovery under diverse challenging conditions, demonstrating strong environmental adaptability and motion robustness, while maintaining full post-recovery motion capability. The framework also generalizes effectively to a high-DOF humanoid, confirming its practical generalizability. The project page is available at \url{https://2350575870.github.io/force-guided.github.io/}.

\end{abstract}

\begin{IEEEkeywords}
Bipedal-Wheeled Robots, Force-Guided Learning, Constrained Policy Optimization, Stage-Wise Reward.
\end{IEEEkeywords}

\section{Introduction}
\IEEEPARstart{B}{ipedal-wheeled} robots combine legged locomotion's terrain adaptability with wheeled mobility's efficiency, drawing significant research 
interest \cite{kashiri_centauro_2019,peng_coordinated_2020}. While some significant progress has been achieved in legged navigation on complex terrain \cite{wang_design_2025,zhao_compliant_2024,chamorro_reinforcement_2024,klemm_ascento_2019,yang_multi-loco_2025}, legged robots remain 
vulnerable to falls under real world uncertainties, often resulting in 
arbitrary, unpredictable postures.

The fall recovery for legged robots has been historically studied through model-based strategies. Early approaches, applicable across various robotic forms, relied on executing fixed, precomputed trajectories often derived from motion capture or trajectory optimization \cite{kanehiro_first_2003}. While offering a deterministic solution, such methods are highly sensitive to the initial pose and lack the robustness and generalization needed for the diverse falling configurations encountered in practice.

To address these limitations, research has pivoted towards data-driven frameworks using reinforcement learning (RL) and imitation learning. Some approaches \cite{li_balance_2025} initiated training from simplified initial poses but replaced model-based optimization with end-to-end policy learning, substantially improving both the success rate and motion robustness of the recovery process. Relevant research \cite{araki_standing-up_2018, tao_learning_2022} advanced to address the more challenging scenarios of recovery from multiple poses and across varied terrains. To date, some studies \cite{hwangbo_learning_2019, lee_robust_2019, gaspard_frasa_2025} often employ staged training frameworks or incorporate curriculum learning, which decomposes the complex recovery task into progressively difficult phases to facilitate stable policy acquisition. Notably, HoST \cite{huang_learning_2025} defines multiple categories of rewards with distinct weighting coefficients to form a composite learning objective while using a force curriculum that applies and gradually phases out upward external assistance. This combined approach enhances the training process by refining policy quality and accelerating convergence. In parallel, imitation learning offers a complementary pathway for generating complex motor skills, enabling robots to acquire more human-like and natural recovery motions by leveraging expert demonstrations or video data \cite{liao_beyondmimic_2025, allshire_visual_2025, he_omnih2o_2024}.

\begin{figure}[t]
   \centering
   \includegraphics[scale=0.43]{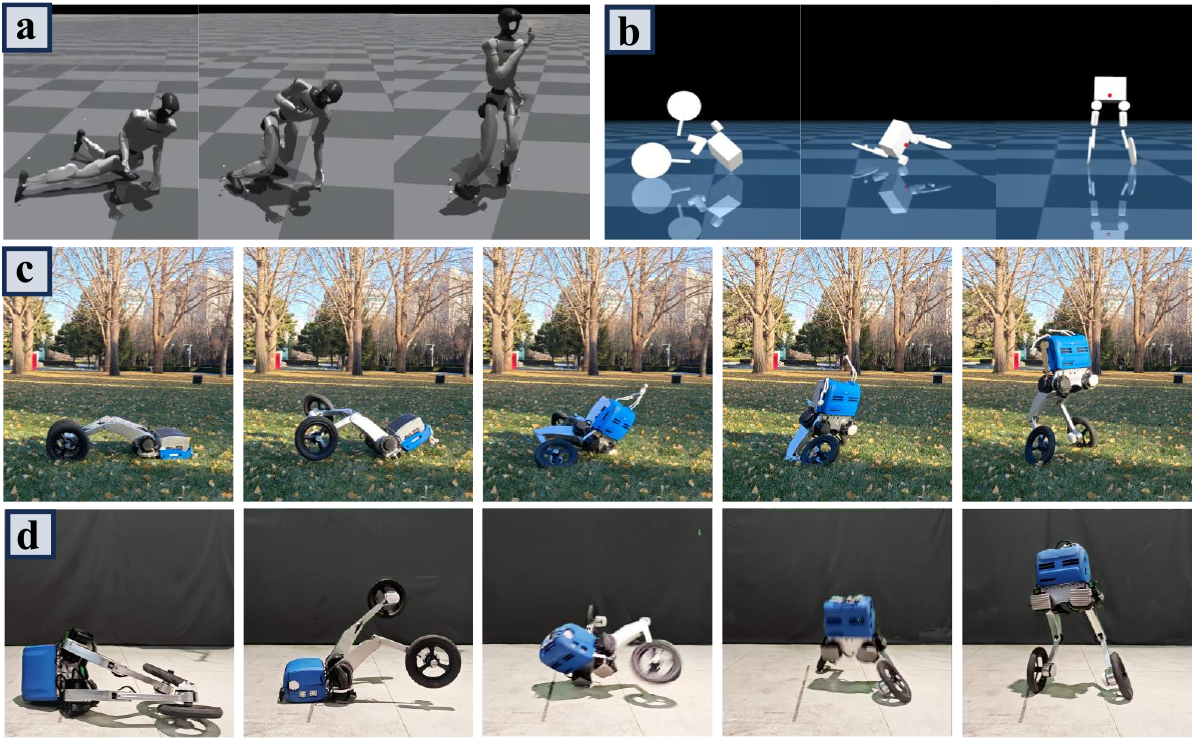}
   \caption{\textbf{Simulation and physical deployment results.} Subfigures (a) and (b) show simulated recovery of a high-DOF Unitree robot and a bipedal-wheeled robot, respectively, while (c) and (d) present successful real-world recovery of the latter under varying environments and initial poses.}
   \label{fig:the first picture in first page}
\end{figure}

In contrast, armless bipedal-wheeled robots face significant challenges in recovery, as they must rely solely on their legs for actuation without the support of arms or additional legs. A dynamic recovery from a specific posture has been demonstrated on bipedal-wheeled robots in non-peer-reviewed video content \cite{LimXDynamics_BV1iSnxzqEd3}. However, existing methods, which are fundamentally reward-based, often converge prematurely to local optima when applied to armless bipedal-wheeled robots. This issue is particularly acute in complex terrains where special circumstances prevent the robots from shifting to a specific initial posture, reducing recovery success rates or causing complete failure. These methods, when training constraints become excessive and convergence conditions overly demanding, often converge prematurely to local optima, termed dead points, resulting in policy training failure. This problem is exacerbated on bipedal-wheeled robots where limited joint actuation is concentrated in the lower body without upper-body support, making premature convergence severely undermine both learning efficiency and real-world robustness.

To address these issues, we propose the FTSR framework. Its core force‑guided learning constructs an external auxiliary force during simulation that correlates directly with the robot's real‑time height, explicitly formulated as optimizable hard constraints \cite{Joshua2017Constrained} rather than a decaying curriculum. This constraint‑driven approach prevents the policy from converging directly to useless local optima, enabling robust and adaptive fall recovery for armless bipedal‑wheeled robots. To achieve stable posture refinement and enable subsequent locomotion after fall recovery, we introduce height-progressive stage-wise rewards that progressively update the target base height based on height thresholds. This key design element ensures stable training, while the stage-wise structure \cite{Kim2025Stage-Wise} guides the robot through posture refinement and subsequent locomotion after standing up. Supplemented by teacher‑student architecture \cite{kim2024NotOnlyRewards} adapted to capture privileged force parameters, FTSR achieves efficient autonomous recovery with real‑world validation, and shows transferability potential to high‑DOF humanoid robots, as shown in Fig. \ref{fig:the first picture in first page}.

\par The main contributions of this work are as follows:

\begin{itemize}
\item The force-guided method we propose constructing height-correlated external auxiliary forces as optimizable constraints is proposed to restrict actions to physically feasible recovery trajectories and enable robust fall recovery for armless bipedal-wheeled robots without supportive assistance, readily extending to high-DOF humanoids.
\item Height-progressive stage-wise rewards enable stable training while guiding the robot through posture refinement and subsequent locomotion after recovery.
\item  For the first time, fall recovery for an armless bipedal-wheeled robot from multiple randomized postures and diverse terrains has been achieved in real-world scenarios, validating the robustness and adaptability of the approach.
\end{itemize}

\section{Method}

To address premature convergence to dead points in armless bipedal-wheeled robot fall recovery, we propose the Force-guided Teacher-student framework with Stage-wise Rewards (Sec. \ref{framework}, Alg. \ref{alg: force-guidance}). Unlike curriculum-based methods that heuristically phase out external assistance, we introduce force-guided learning that formulates auxiliary forces and torques as optimizable constraints within CPO (Sec. \ref{CMDP}), guiding the policy toward reducing force dependency gradually as it develops recovery strategies without arm support (Sec. \ref{force-guided}). Meanwhile, height-progressive stage-wise rewards automatically transition from posture refinement to locomotion based on batch height statistics, ensuring stable training (Sec. \ref{reward-design}).

\subsection{Constrained Markov Decision Process} 
\label{CMDP}
A Constrained Markov Decision Process (CMDP) \cite{altman2021constrained} is an extension of the Markov Decision 
Process, which must satisfy additional constraints (such as resource limitations or 
safety requirements) while optimizing the long-term cumulative reward. In this paper, 
we employ a CMDP to achieve progressive policy training from strong guidance to autonomy 
through tightenable external-force constraints. A CMDP can be formalized as a tuple 
$\langle S,A,P,R,\gamma,C,d \rangle$, where $S$ denotes the state space, $A$ is action space, $P$ is state transition 
probability, $R\left(s_{t}, a_{t}, s_{t+1}\right)$ denotes the immediate reward function, $\gamma$ is discount factor, 
$C={C_i(s_t,a_t,s_{t+1})}_{i=1}^m$ represents $m$ constraint costs incurred when executing an action 
(each constraint function $C_i:S\times A\times S\rightarrow\mathbb{R}$ is defined similarly 
to the reward function), and $d=\{d_1,\cdots, d_i,\cdots,d_m\}$ denotes the allowable 
upper limits for each constraint cost. Our objective is to learn an optimal policy 
$\pi^\ast:S\rightarrow A$ that maximizes the expected discounted return while satisfying 
all constraints over the trajectory $\tau$:
\begin{equation}
  \begin{gathered}
    J(\pi)={\mathbb{E}}_{\pi}\left[\sum_{t=0}^{\infty} \gamma^{t} R\left(s_{t}, a_{t}, s_{t+1}\right)\right] \\
    \text{s.t. } J_{C_{i}}(\pi) \leq d_{i} \quad \forall i \in\{1, \ldots, m\}
  \end{gathered}
\end{equation}

\noindent In the above formulation,
$J_{C_i}(\pi)$ represent the expected 
return of the $i$-th constraint under policy $\pi$, respectively, defined as follows:


\begin{equation}
  J_{C_{i}}(\pi)={\mathbb{E}}_{\pi}\left[\sum_{t=0}^{\infty} \gamma^{t} C_{i}\left(s_{t}, a_{t}, s_{t+1}\right)\right]
\end{equation}

To solve the aforementioned optimization problem, \cite{Joshua2017Constrained} and \cite{schulman2015trust} 
proposed the Constrained Policy Optimization (CPO) algorithm, which integrates 
constraints into the trust-region framework to enable larger update steps:

\begin{equation}
\label{cpo-func}
\begin{gathered}
\pi_{k+1}=\arg \max _{\substack{\pi \in \Pi_{\theta}}} \underset{\substack{s \sim d^{\pi_{k}} \\ a \sim \pi}}{\mathbb{E}}\left[A^{\pi_{k}}(s, a)\right] \\
\text { s.t. } J_{C_{i}}\left(\pi_{k}\right)+\frac{1}{1-\gamma} \underset{\substack{s \sim d^{\pi_{k}} \\ a \sim \pi}}{\mathbb{E}}\left[A_{C_{i}}^{\pi_{k}}(s, a)\right] \leq d_{i} \quad \forall i \\
\bar{D}_{K L}\left(\pi \| \pi_{k}\right) \leq \delta
\end{gathered}
\end{equation}

\noindent where ${\Pi_{\theta}}$ is the set of parameterized policies with parameters $\theta$, $d^{\pi}$ denotes the 
state visitation distribution under policy $\pi$, defined by $d^{\pi}(s)=(1-\gamma) \sum_{t=0}^{\infty} \gamma^{t} P\left(s_{t}=s \mid \pi\right)$,
$\bar{D}_{K L}\left(\pi \| \pi_{k}\right)=\mathbb{E}_{s \sim d^{\pi_{k}}} \left[D_{K L}\left(\pi \| \pi_{k}\right)[s]\right]$ 
represents the expected KL-divergence between policies, $\delta>0$ is the maximum step 
size, $A^{\pi_k}(s,a)$ is the advantage function for the reward, and $A_{C_i}^{\pi_k}(s,a)$ is the advantage function for the $i$-th constraint.

\subsection{Force-Guided Reinforcement Learning Framework}
\label{framework}
\begin{figure*}[t] 
  \centering
  \includegraphics[width=\textwidth]{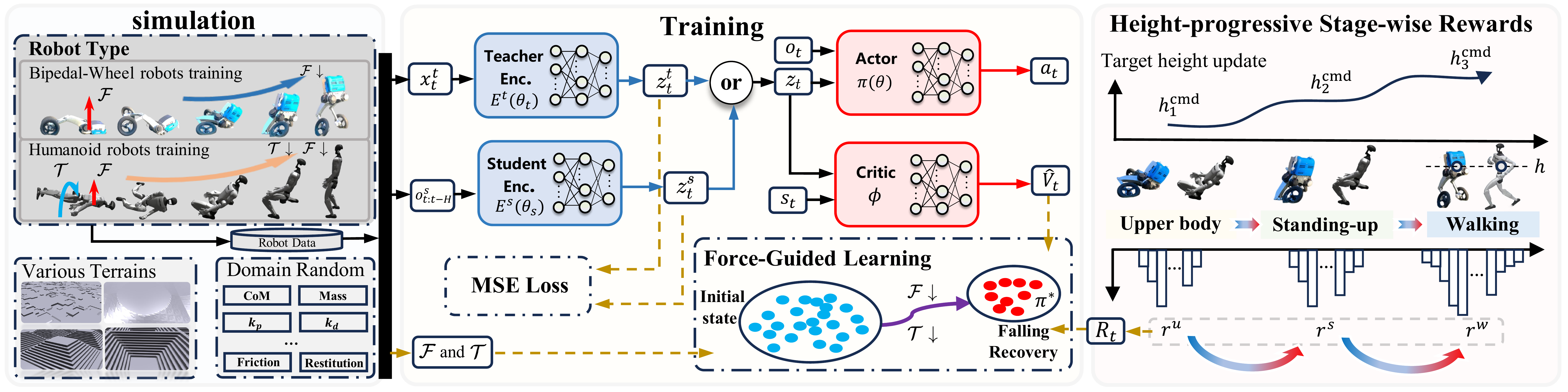} 
  \caption{\textbf{Force‑guided fall‑recovery training framework.} The framework supports either bipedal-wheeled or humanoid robots, selected from 'Robot Type'. Training data flows from the selected robot through "Robot Data" to generate teacher encoder observation ${x}^t_t$ and student history observation ${o}^s_{t:t-H}$, which feed $E^t$ and $E^s$ respectively to produce latents ${z}^t_t$ and ${z}^s_t$. These latents are then selected by an "or" module, with ${z}^t_t$ for teacher-group agents and ${z}^s_t$ for student-group agents, to form ${z}_t$ concatenated with ${o}_t$ and ${s}_t$ for shared Actor-Critic networks ($\pi$ and $\phi$) outputting ${a}_t$ and $\hat{V}_t$. During training, yellow arrows indicate training signals: ${z}^t_t$ and ${z}^s_t$ supervise $E^s$ via MSE loss, while $\mathcal{F}$, $\mathcal{T}$, $R_t$, and $\hat{V}_t$ update $E^t$, $\pi$, and $\phi$ to evolve policy toward $\pi^*$ by progressively reducing $\mathcal{F}$ and $\mathcal{T}$ via Force-Guided Learning. As robot base height increases, $R_t$ transitions from $r^u$ to $r^w$ when agents exceeding height thresholds reach two-thirds of the agents, synchronizing target height and rewards with actual height progression.}
  \label{fig:framework argitacture}
\end{figure*}

The overall framework is illustrated in Fig.~\ref{fig:framework argitacture}. Armless bipedal-wheeled robots lack upper-body support during fall recovery, making them prone to convergence to dead points under pure reward optimization. This motivates formulating external auxiliary forces $\mathcal{F}$ and torques $\mathcal{T}$ as optimizable constraints rather than decaying curricula, enabling the policy to gradually reduce dependency on external assistance while discovering physically feasible recovery trajectories. Such constraint-guided exploration necessitates stable posture refinement across phases, which we achieve through height-progressive stage-wise rewards that auto-transition based on batch height statistics rather than fixed durations.

Realizing these mechanisms requires balancing privileged information with proprioceptive-only execution, leading to a teacher-student architecture where teacher encoder $E^t$ processes ${x}^t_t$ comprising contact forces, base height, and full bodies state, while student encoder $E^s$ learns equivalent representations from history ${o}^s_{t:t-H}$. This history concatenates observations ${o}_t$ over past $H$ steps, where each ${o}_t \in \mathbb{R}^{34}$ includes angular velocity, gravity projection, user commands, joint positions, velocities, and previous actions. An "or" module routes latent ${z}^t_t$ or ${z}^s_t$ to shared Actor-Critic networks by agent group, producing ${z}_t$ concatenated with current ${o}_t$ and ${s}_t \in \mathbb{R}^{188}$ containing height map and robot base height. The networks output actions ${a}_t$ and values $\hat{V}_t$, with parameters in Table~\ref{tab:Network Architecture} and component designs in Sec.~\ref{force-guided} and Sec.~\ref{reward-design}.

\subsection{Force-Guided Fall Recovery Training}
\label{force-guided}
Force curriculum heuristically phases out assistance, causing collapse before stable standing is acquired and convergence to dead points under excessive constraints. We instead propose force-guided learning, formulating forces and torques as optimizable constraints within CPO, reshaping their functional forms to enable gradual reduction tied to policy improvement. The expressions for force $\mathcal{F}$ and torque $\mathcal{T}$ are given in Equation (\ref{force-torque-func}).

\begin{equation}
\label{force-torque-func}
\begin{aligned}
\left[\begin{array}{l}
\mathcal{F} \\
\mathcal{T}
\end{array}\right] &= \left(1-e^{-\mu\left(h^{\text{cmd}}_i-h\right)}\right) \operatorname{sat}_{[0,1]}\left(t_{\text{coeff}}\right) \\
&\quad \times \left[\begin{array}{c}
F_{\max} n \\
T_{\max} \log \left(R_{\text{tag}} R^{-1}\right)
\end{array}\right]
\end{aligned}
\end{equation}
\noindent In the formulation, $\mu$ is the height coefficient, which determines 
the extent to which height influences the force and torque. 
$h^{\text{cmd}}_i$ denotes the target height of the center of mass (CoM), 
while $h$ represents the current CoM height. The time coefficient is 
given by 
$t_{\text {coeff }}=1-t / t_{\text {tag }}^{\text {step }}$, where $t$ is the 
current time step and $t_{\text{tag}}^{\text{step}}$ is the step at which 
external assistance ends, 
and $\text{sat}_{[0,1]}\left( \cdot \right)$ denotes a saturation function that clips the parameter 
to the range $[0,1]$. $F_{\max}$ indicates the maximum external force 
applied to the robot, and $T_{\max}$ is the maximum applied torque. 
${R}_{\text{tag}}$ is the target rotation matrix generated 
from the target Euler angles, and ${R}$ is the rotation matrix 
derived from the current Euler angles. The force axis is defined by 
the vector $n = \begingroup\setlength{\arraycolsep}{2pt}\begin{bmatrix} 0 & 0 & 1 \end{bmatrix}^T\endgroup$
. The 
notation $\log(\cdot)$ represents the logarithmic map on the 
$so(3)$ group, which maps the relative rotation matrix 
${R}_{\text{tag}}{R}^{-1}$ to an instantaneous rotation 
vector $\mathbf{w}\in\mathbb{R}^3: \mathbf{w}=\log({R}_{\text{tag}}{R}^{-1})$. 
The direction of this vector corresponds to the axis of rotation, and 
its 2-norm equals the rotation angle in radians.

\begin{table}[t]
\centering
\caption{Network Architecture}
\label{tab:Network Architecture}
\renewcommand{\arraystretch}{1.2}
\begin{tabular}{llll}
\toprule

\textbf{Models} & \textbf{Input} & \textbf{Hidden layer} & \textbf{Output} \\
\midrule

Actor ${\pi}({\theta})$ & ${z}_t,{{o}}_{t}$ & $\displaystyle [512,256,128]$ & $\displaystyle {a}_t$ \\
Critic ${\phi}$ & ${z}_t,{{s}}_t$ & $\displaystyle [512,256,128]$ & $\displaystyle {\hat{V}}_t$ \\

Privileged Enc. $E^{t}(\theta_t)$ & ${{x}}_t^t$ & $\displaystyle [256,128]$ & $\displaystyle {z}^{{t}}_t$ \\
Student Enc. $E^{s}(\theta_s)$ & ${o}_{t:t-H}^s$ & $\displaystyle [256,128]$ & $\displaystyle {z}^{s}_t$ \\
\bottomrule
\end{tabular}
\end{table}

However, the goal is not merely to apply forces, but to train a policy that adapts to them, learning recovery with progressively reduced intervention. Inspired by CPO, 
we formulate the force and torque as two constraint conditions: 
$C_1=\mathcal{F}$ with $J_{C_1}(\pi)\leq d_1$ and $C_2=\mathcal{T}$ with $J_{C_2}(\pi)\leq d_2$, 
and integrate them into the training process. We then incorporate these two constraints into Equation (\ref{cpo-func}). Unlike the original CPO formulation that solves constrained quadratic programs, we employ the penalty function method to transform the optimization into a simpler form containing only a KL-divergence:

\begin{equation}
\label{simper-form-func}
\renewcommand{\jot}{-1pt}
\begin{gathered}
\underset{\pi \in \Pi_{\theta}}{\arg \max} \; \mathop{\mathbb{E}}_{\substack{s \sim d^{\pi_{k}} \\ a \sim \pi}}\left[A^{\pi_{k}}(s, a)\right] \\
+\sum_{i=1}^{m} \beta_{i}\left(d_{i}-\left(J_{C_{i}}\left(\pi_{k}\right)+\frac{1}{1-\gamma} \mathop{\mathbb{E}}_{\substack{s \sim d^{\pi_{k}} \\ a \sim \pi}}\left[A_{C_{i}}^{\pi_{k}}(s, a)\right]\right)\right) \\
\text{s.t. } \bar{D}_{K L}\left(\pi \| \pi_{k}\right) \leq \delta
\end{gathered}
\end{equation}

\noindent  In the formulation, $A^{\pi_k}(s,a)$ and $A_{C_i}^{\pi_k}(s,a)$ are computed using 
Generalized Advantage Estimation (GAE). 
Prior to gradient calculation using the advantage functions, an 
additional standardization step is applied to enhance training 
stability. Equation (\ref{simper-form-func}) is optimized via Proximal Policy Optimization 
(PPO). 

To enable the policy model to directly access privileged information, 
thereby accelerating convergence and improving training stability, 
we employ a teacher-student model \cite{Wang2024CTS}. Its objective 
function is given by Equation (\ref{ppo-func}):

\begin{equation}
\label{ppo-func}
\begin{aligned}
L^{\text{cpo},j}(\vartheta) &= \dfrac{1}{\left|\mathcal{D}^{j}T\right|} \sum_{\tau \in \mathcal{D}^{j}} \sum_{t=0}^{T} \\
&\quad \min\left(r_{t}^{j} \bar{A}_{t}^{j}, \text{sat}_{[1-\epsilon,1+\epsilon]}\left(r_{t}^{j}\right) \bar{A}_{t}^{j}\right)
\end{aligned}
\end{equation}

\noindent In this formulation, the superscript $(\cdot)^j$ denotes whether the teacher or 
the student model is being used ($j=t$ indicates teacher-group parameters, 
and $j=s$ indicates student-group parameters). The subscript $(\cdot)_t$ 
denotes the value of the bracketed parameter at time step $t$. $\vartheta$ represents 
the mixed network weights: for the teacher group, 
$\vartheta=\theta+\theta^t$; for the student group, $\vartheta=\theta$. 
$T$ is the trajectory length of the agent-environment interaction, 
and $\mathcal{D}^j$ is the trajectory data generated by the corresponding policy. 
$r_t^j$ is the importance sampling ratio, defined explicitly in Equation 
(\ref{ratio-func}). $\epsilon$ is the clipping range parameter in PPO. $\bar{A}_{t}^{j}$ denotes the 
mixed advantage function that combines the value function and 
constraint functions, as defined in Equation (\ref{advantage-func}).

\begin{equation}
\label{ratio-func}
r_{t}^{j} = \left\{
\begin{array}{ll}
\dfrac{\pi\left(a_{t}^{t} \mid o_{t}, E^{t}\left(x_{t}^t\right)\right)}{\pi_{\text{old}}\left(a_{t}^{t} \mid o_{t}, E^{t}\left(x_{t}^t\right)\right)} & j=t \\[12pt]
\dfrac{\pi\left(a_{t}^{s} \mid o_{t}, E^{s}\left(o_{t: t-H}^s\right)\right)}{\pi_{\text{old}}\left(a_{t}^{s} \mid o_{t}, E^{s}\left(o_{t: t-H}^s\right)\right)} & j=s
\end{array}
\right.
\end{equation}

\begin{equation}
\label{advantage-func}
\begin{split}
\bar{A}_{t}^{j} &= A_{t}^{j}(s, a) - \sum_{i=1}^{m} \beta_{i} \\
&\quad \left( J_{t, C_{i}}^{j}(\pi_{k}) + \frac{1}{1-\gamma} \underset{\substack{s \sim d^{\pi_{k}} \\ a \sim \pi}}{\mathbb{E}}\left[A_{t, C_{i}}^{j}(s, a)\right] \right)
\end{split}
\end{equation}

\noindent In the formulation, $J_{t,C_i}^j(\pi_k)$ denotes the constraint function 
value at time step $t$ in the trajectory, and $A_{t,C_i}^j(s,a)$ is the 
corresponding constraint advantage function. Unlike Equation (\ref{simper-form-func}), our 
training objective aims to drive the external forces and torques acting 
on the agent to zero; hence, $d_i$ should be as small as possible. 
To ensure the mixed advantage function simplifies to the target form $\bar{A}_t^j = A_t^j(s, a)$ as given in Equation (\ref{advantage-func}), we set $d_i = 0$ when $t = t_{\text {tag }}^{\text {step }}$.

The Critic network is trained using a mixed dataset $\mathcal{D}=\mathcal{D}^t\cup \mathcal{D}^s$ 
and updated by minimizing the mean squared error between its predicted 
value $\hat{V_t}$ and the target value $R_t$ estimated from trajectory 
returns:

\begin{equation}
  L^{\text {value }}(\phi)=\frac{1}{\left|\mathcal{D} T \right|} \sum_{\tau \in \mathcal{D}} \sum_{t=0}^{T}\left(\hat{V}_{t}-R_{t}\right)^{2}
\end{equation}

For the student encoder, we employ the Mean Squared Error (MSE) as the 
training objective. Unlike the approach in \cite{Wang2024CTS}, we do not 
restrict training to trajectories generated solely by the student 
group; instead, we directly utilize all trajectories obtained from 
agent-environment interactions for training.

\begin{equation}
  L^{s}\left(\theta_{s}\right)=\frac{1}{\left|\mathcal{D} T \right|} \sum_{\tau \in \mathcal{D}} \sum_{t=0}^{T}\left\|E^{s}\left(o^s_{t: t-H}\right)-E^{t}\left(x^t_{t}\right)\right\|_{2}^{2}
\end{equation}

\begin{algorithm}
  \caption{FTSR Algorithm}
  \label{alg: force-guidance}
  \begin{algorithmic}[1]
    \State Initialize environment and networks
    \State Initialize Penalty Factor $\beta_{i}$, Learning Rate $\alpha_{cpo}, \alpha_{mse}$ for training.
    \State Empty replay buffer $\mathcal{D}^t$ and $\mathcal{D}^s$
    \State Use $\vartheta$ to denote $\theta$ and $\theta_t$ for brevity

    \For{$0 \leq \text{its} \leq \text{iterations}$}
        \State Compute $\mathcal{F}$ and $\mathcal{T}$ using Eq.~(\ref{force-torque-func})
        \State $C_i \gets \left( \mathcal{F},\ \mathcal{T} \right)$
        \State Compute $A_{t}^j$ using GAE, compute $\bar{A}_t^j$ using Eq.~(\ref{advantage-func})
        \State {Update reward via threshold vector:}
        \State $R_t \gets [r^u, r^w, r^s] \cdot [\mathbb{I}(p=0), \mathbb{I}(p=1), \mathbb{I}(p=2)]^\top$
        \State \quad where $p = 2 \cdot \mathbb{I}(|S_1| > \frac{2}{3}N) + \mathbb{I}(|S_2| > \frac{2}{3}N)$

        \State Actor-Critic and Teacher Enc.\ update:
        \State $\vartheta \gets \vartheta + \alpha_{cpo}\nabla_{\vartheta}
                 (L^{\text{cpo},t}(\vartheta) +
                  L^{\text{cpo},s}(\theta))$
        \State $\phi \gets \phi + \alpha_{cpo}\nabla_{\phi}
                 L^{\text{value}}(\phi)$
    \EndFor
    \For{epoch $i = 0,1,\dots$}
        \State Student Enc.\ update using MSE:
        \State $\theta_s \gets \theta_s +
                 \alpha_{mse}\nabla_{\theta_s}
                 (L^{s}(\theta_s)$)
    \EndFor
  \end{algorithmic}
\end{algorithm}

\subsection{Height-progressive Stage-wise Rewards}
\label{reward-design}
To address training instability caused by fixed target heights in prior stage-wise reward designs, we adopt height-progressive stage-wise rewards that progressively update target base heights based on agents-level height thresholds. The final reward functions and their weights are listed in Table \ref{tab:reward terms and weights}. Each layer employs a distinct reward set to guide the robot sequentially toward recovery: \textbf{(1) Upper Body Erection $r^u$.} This initial phase aims to drive the robot’s upper body from a prone position to an upright posture, primarily guided by a gravity projection penalty, with the target base height set to $h_1^{\text{cmd}}$. We define agent sets $S_1 = \{ i \mid h > h_1^{\text{cmd}} \}$ and $S_2 = \{ i \mid h > h_2^{\text{cmd}} \}$ based on height thresholds. \textbf{(2) Standing-up $r^s$.} This phase is activated when $|S_1| > \frac{2}{3}N$, i.e., when the number of agents whose base height exceeds $h_1^{\text{cmd}}$ reaches two-thirds of the total population $N$. The objective is to encourage the robot to retract its lower body, bringing the leg joints back to angles close to the initial standing configuration to achieve a stable standing pose, and the target base height is updated to $h_2^{\text{cmd}}$. \textbf{(3) Walking $r^w$.} The final phase begins when $|S_2| > \frac{2}{3}N$, meaning the number of agents with $h > h_2^{\text{cmd}}$ reaches two-thirds of the total. This phase promotes normal locomotion after standing up, preventing the robot from remaining stationary, and the target base height is further updated to $h_3^{\text{cmd}}$. The symbol "-" denotes that the reward is inactive for that stage.

While the primary reward functions are adopted from \cite{Wang2024CTS}, we specifically designed several additional reward terms (highlighted in red in Table \ref{tab:reward terms and weights}) to address key challenges:
\begin{itemize}

\item \textbf{Termination Penalty:} To prevent agents from lying idle on the ground, any agent contacting the ground for over 10 seconds is terminated, and a penalty is applied based on the number of terminated agents $n^{\text{te}}$.

\item \textbf{Posture Penalties:} To penalize unnatural postures, we penalize aerial motion by counting agents with both wheels off the ground $n^{\text{nf}}$, penalize changes in leg spacing using the Y-coordinates of the left and right legs in the robot frame $p_y^{\text{lw}}$, $p_y^{\text{rw}}$, and penalize asymmetric leg poses by calculating the difference between left and right leg joint angles $q^{\text{left}}-q^{\text{right}}$.

\item \textbf{Leg-Usage Reward:} o encourage leg usage during standing, we reward agents when wheel contact force $f^{\text{wheel}}$ approaches total weight $M$.

\end{itemize}

\begin{table}[t]
\centering
\caption{Reward Terms and Weights}
\label{tab:reward terms and weights}
\renewcommand{\arraystretch}{1.1}
\footnotesize
\setlength{\tabcolsep}{2.5pt} 
\begin{tabularx}{\linewidth}{@{}>{\raggedright}p{1.4cm}>{\raggedright}Xccc@{}}
\toprule
\textbf{Reward} & \textbf{Definition} & \textbf{\makebox[0.5cm][c]{$r^u$}} & \textbf{\makebox[0.5cm][c]{$r^s$}} & \textbf{\makebox[0.5cm][c]{$r^w$}} \\
\midrule
Lin. vel. & $\exp(-8.3\| {v}_{xy} - {v}_{xy}^{\text{cmd}} \|_2^2)$ & - & - & 7.0 \\
Ang. vel. & $\exp(-8.3\| {\omega}_z - {\omega}_z^{\text{cmd}} \|_2^2)$ & - & - & 4.0 \\
Orient. & $\| \mathbf{g}_{xy} \|_2$ & -2.6 & -2.8 & -10.0 \\
Torques & $\| {\tau} \|_2^2$ & $-1.0e^{-6}$ & $-1.0e^{-6}$ & $-1.0e^{-6}$ \\
DOF acc. & $\ddot{{q}}^2$ & $-2.5e^{-8}$ & $-2.5e^{-6}$ & $-2.5e^{-6}$ \\
DOF vel. & $\dot{{q}}^2$ & $-$ & $-$ & $-0.01$ \\
Act. rate & $\| {a}_t - {a}_{t-1} \|_2^2$ & -0.06 & -0.02 & -0.08 \\
Act. smooth & $\| {a}_t - 2{a}_{t-1} + {a}_{t-2} \|_2^2$ & - & - & -0.04 \\
Dof pos. & $\| {q} - {q}_{\text{norminal}} \|_2^2$ & -0.06 & -0.1 & -0.12 \\
Dof ener. & $\| \dot{{q}}{\tau} \|_2^2$ & - & - & $-1.0e^{-4}$ \\

\textcolor{red}{Base heig.} & \textcolor{red}{$\exp(-8.3\| {h} - {h}^{{\text{cmd}}}_i \|_2^2)$} & \textcolor{red}{4.0} & \textcolor{red}{6.0} & \textcolor{red}{5.0} \\
\textcolor{red}{Termin.} & \textcolor{red}{$n^{\text{te}}$} & \textcolor{red}{-200} & \textcolor{red}{-100} & \textcolor{red}{-500} \\
\textcolor{red}{Feet dist.} & \textcolor{red}{$\| |{p}_y^{\text{lw}} - {p}_y^{\text{rw}}| - 0.35 \|_2$} & \textcolor{red}{-} & \textcolor{red}{-} & \textcolor{red}{-2.0} \\
\textcolor{red}{Leg bias} & \textcolor{red}{$\| {q}^{\text{left}} - {q}^{\text{right}} \|_2$} & \textcolor{red}{-2.0} & \textcolor{red}{-0.02} & \textcolor{red}{-} \\
\textcolor{red}{No fly} & \textcolor{red}{$n^{\text{nf}}$} & \textcolor{red}{-2.0} & \textcolor{red}{-} & \textcolor{red}{-0.2} \\
\textcolor{red}{Wheel force} & \textcolor{red}{$\| f^{\text{wheel}}-M \|_2^2$} & \textcolor{red}{4.0} & \textcolor{red}{-} & \textcolor{red}{-} \\
\bottomrule
\end{tabularx}
\end{table}

\section{Experiments and Results}

All training was conducted in IsaacGym \cite{makoviychuk2021isaac} on an NVIDIA GeForce RTX 4070 Super GPU. We simulated 4,000 parallel environments (with 3,000 teacher and 1,000 student agents) for 8,000 iterations. The policy was trained at 200 Hz and updated at 50 Hz, while the hardware operated at 500 Hz to balance training speed and stability.

\subsection{Training Details}
\textbf{Model Initialization:} Our method embeds the penalty-function approach within a global stochastic search framework, which typically requires extended training. Since RL performance often degrades in long-horizon tasks, direct training from random initialization may fail to converge. Rather than deepening the network, which can yield low-success-rate policies, we initialize training from a basic model pre-trained with $r^w$. Training stops once the robot achieves elementary walking motion, usually within about 200 iterations.

\textbf{Terrain:} To improve robustness and sim-to-real transfer, training is conducted across five terrain types: flat ground, slopes ($\leq 20^\circ$), pyramid steps (step height $\leq 0.12\text{m}$), discrete uneven terrain (height variation $\leq 0.12\text{m}$), and rough surfaces (roughness $\leq 0.03\text{m}$). This diversity encourages the policy to generalize across environments and helps bridge the sim-to-real gap induced by model discrepancies.

\begin{table}[t]
\centering
\caption{Domain Randomization}
\label{tab:domain_randomization}
\renewcommand{\arraystretch}{1.1}
\begin{tabular}{llll}
\toprule
\textbf{Parameter} & \textbf{Randomization Range} & \textbf{Unit} \\
\midrule
Base Mass & $\mathcal{U}(-0.1, 1.2)$ & kg \\
Base $\text{CoM}_x$ offset & $\mathcal{U}(-0.02,0.02)$ & m \\
Base $\text{CoM}_y$ offset & $\mathcal{U}(-0.01,0.01)$ & m \\
Base $\text{CoM}_z$ offset & $\mathcal{U}(-0.02,0.02)$ & m \\
Friction & $\mathcal{U}(0.1, 1.7)$ & - \\
Restitution & $\mathcal{U}(0.3, 1.0)$ & - \\
$k_{\text{p}}$ and $k_{\text{d}}$ & $\mathcal{U}(0.95,1.05) \times \text{default}$ & Nm/rad \\
Motor Strength & $\mathcal{U}(0.85,1.05)$ & - \\
Initial joint angle & $\mathcal{U}(0.5,1.5) \times \text{default joint angle}$ & - \\
Initial base euler & $\mathcal{U}(-0.3,0.3)$ & - \\
\bottomrule
\end{tabular}
\end{table}

\textbf{Domain Randomization:} To bridge the sim-to-real gap, we applied domain randomization to key parameters during training (Table \ref{tab:domain_randomization}), including PD gains, center-of-mass offset, torque offset, and initial offsets for joint position/velocity and base pose. The robot’s initial base pose was randomized across four states—front, down, or lying on either side—with an additional pose offset applied as perturbation.

\begin{table*}[htbp]
    \centering
    \caption{Ablation Study Results (mean$\pm$std).}
    \label{tab:wheel_robot_sim_tab}
    \resizebox{\textwidth}{!}{
    \setlength{\tabcolsep}{4.5pt} 
    \renewcommand{\arraystretch}{1.2} 
    \scriptsize
    \begin{tabular}{p{2.2cm}cccccccccccc}
        \toprule
        Method & \multicolumn{4}{c}{Ground} & \multicolumn{4}{c}{Slope} & \multicolumn{4}{c}{Step} \\
        \cmidrule(lr){2-5} \cmidrule(lr){6-9} \cmidrule(lr){10-13}
        & $E_{\text{succ}}$ $\uparrow$ & $E_{\text{feet}}$ $\downarrow$ & $E_{\text{smth}}$ $\downarrow$ & $E_{\text{engy}}$ $\downarrow$ & $E_{\text{succ}}$ $\uparrow$ & $E_{\text{feet}}$ $\downarrow$ & $E_{\text{smth}}$ $\downarrow$ & $E_{\text{engy}}$ $\downarrow$ & $E_{\text{succ}}$ $\uparrow$ & $E_{\text{feet}}$ $\downarrow$ & $E_{\text{smth}}$ $\downarrow$ & $E_{\text{engy}}$ $\downarrow$ \\
        \midrule
        
        \addlinespace[-1pt]
        \multicolumn{13}{l}{\textbf{(a) Ablation on  Height-progressive Stage-Wise Rewards}} \\
        \addlinespace[-4pt]
        \multicolumn{13}{c}{\dotfill} \\ 
        Ours w/o SWR & 0.0 & / & / & / & 0.0 & / & / & / & 0.0 & / & / & / \\
        Ours & $\mathbf{99.8_{(\pm 0.7)}}$ & $\mathbf{0.52_{(\pm 0.2)}}$ & $\mathbf{5.80_{(\pm 3.7)}}$ & $\mathbf{4.36_{(\pm 0.2)}}$ & $\mathbf{99.5_{(\pm 0.4)}}$ & $\mathbf{0.68_{(\pm 0.3)}}$ & $\mathbf{7.05_{(\pm 6.19)}}$ & $\mathbf{5.01_{(\pm 0.7)}}$ & $\mathbf{99.8_{(\pm 0.44)}}$ & $\mathbf{0.53_{(\pm 0.1)}}$ & $\mathbf{6.83_{(\pm 2.5)}}$ & $\mathbf{4.12_{(\pm 0.7)}}$ \\
        \addlinespace[2pt]
        
        \cline{1-13}
        \addlinespace[2pt]
        \multicolumn{13}{l}{\textbf{(b) Ablation on Force-Guided}} \\
        \addlinespace[-4pt]
        \multicolumn{13}{c}{\dotfill} \\
        Ours w/ Force Curr. & 0.0 & / & / & / & 0.0 & / & / & / & 0.0 & / & / & / \\
        Ours w/o Force Guid. & $97.4_{(\pm 2.7)}$ & $0.82_{(\pm 1.0)}$ & $12.4_{(\pm 3.5)}$ & $\mathbf{4.1_{(\pm 0.3)}}$ & $98.4_{(\pm 0.4)}$ & $\mathbf{0.42_{(\pm 0.2)}}$ & $12.75_{(\pm 2.84)}$ & $5.69_{(\pm 0.3)}$ & $\mathbf{100.0_{(\pm 0.0)}}$ & $\mathbf{0.3_{(\pm 0.1)}}$ & $12.5_{(\pm 2.8)}$ & $5.05_{(\pm 0.4)}$ \\
        Ours-Force0.02 & $87.8_{(\pm 3.6)}$ & $4.29_{(\pm 8.5)}$ & $12.28_{(\pm 4.8)}$ & $5.53_{(\pm 0.85)}$ & $99.5_{(\pm 0.8)}$ & $0.69_{(\pm 0.1)}$ & $10.77_{(\pm 4.9)}$ & $5.59_{(\pm 0.5)}$ & $99.0_{(\pm 0.4)}$ & $4.67_{(\pm 0.1)}$ & $8.48_{(\pm 6.9)}$ & $5.48_{(\pm 0.3)}$ \\
        Ours & $\mathbf{99.8_{(\pm 0.7)}}$ & $\mathbf{0.52_{(\pm 0.2)}}$ & $\mathbf{5.80_{(\pm 3.7)}}$ & $4.36_{(\pm 0.2)}$ & $\mathbf{99.5_{(\pm 0.4)}}$ & $0.68_{(\pm 0.3)}$ & $\mathbf{7.05_{(\pm 6.19)}}$ & $\mathbf{5.01_{(\pm 0.7)}}$ & $99.8_{(\pm 0.44)}$ & $0.53_{(\pm 0.1)}$ & $\mathbf{6.83_{(\pm 2.5)}}$ & $\mathbf{4.12_{(\pm 0.7)}}$ \\
        \addlinespace[2pt]

        \cline{1-13}
        \addlinespace[2pt]
        \multicolumn{13}{l}{\textbf{(c) Ablation on Teacher-Student}} \\
        \addlinespace[-4pt]
        \multicolumn{13}{c}{\dotfill} \\
        Ours w/o T-S & $59.0_{(\pm 10.7)}$ & $1.5_{(\pm 0.6)}$ & $50.8_{(\pm 8.1)}$ & $5.52_{(\pm 0.5)}$ & 0.0 & / & / & / & 0.0 & / & / & / \\
        Ours w/ RMA Tea. & $\mathbf{100.0_{(\pm 0.0)}}$ & $0.6_{(\pm 0.3)}$ & $\mathbf{4.80_{(\pm 1.7)}}$ & $5.36_{(\pm 0.12)}$ & $\mathbf{99.8_{(\pm 0.7)}}$ & $\mathbf{0.56_{(\pm 0.4)}}$ & $\mathbf{6.12_{(\pm 4.23)}}$ & $\mathbf{4.46_{(\pm 0.7)}}$ & $99.0_{(\pm 0.54)}$ & $0.71_{(\pm 0.1)}$ & $\mathbf{5.71_{(\pm 1.5)}}$ & $6.34_{(\pm 0.8)}$ \\
        Ours & $99.8_{(\pm 0.7)}$ & $\mathbf{0.52_{(\pm 0.2)}}$ & $5.80_{(\pm 3.7)}$ & $\mathbf{4.36_{(\pm 0.2)}}$ & $99.5_{(\pm 0.4)}$ & $0.68_{(\pm 0.3)}$ & $7.05_{(\pm 6.19)}$ & $5.01_{(\pm 0.7)}$ & $\mathbf{99.8_{(\pm 0.44)}}$ & $\mathbf{0.53_{(\pm 0.1)}}$ & $6.83_{(\pm 2.5)}$ & $\mathbf{4.12_{(\pm 0.7)}}$ \\
        \addlinespace[2pt]
        
        \bottomrule
    \end{tabular}}
\end{table*}

\begin{figure}[!t]
\centering
\includegraphics[width=3.45in]{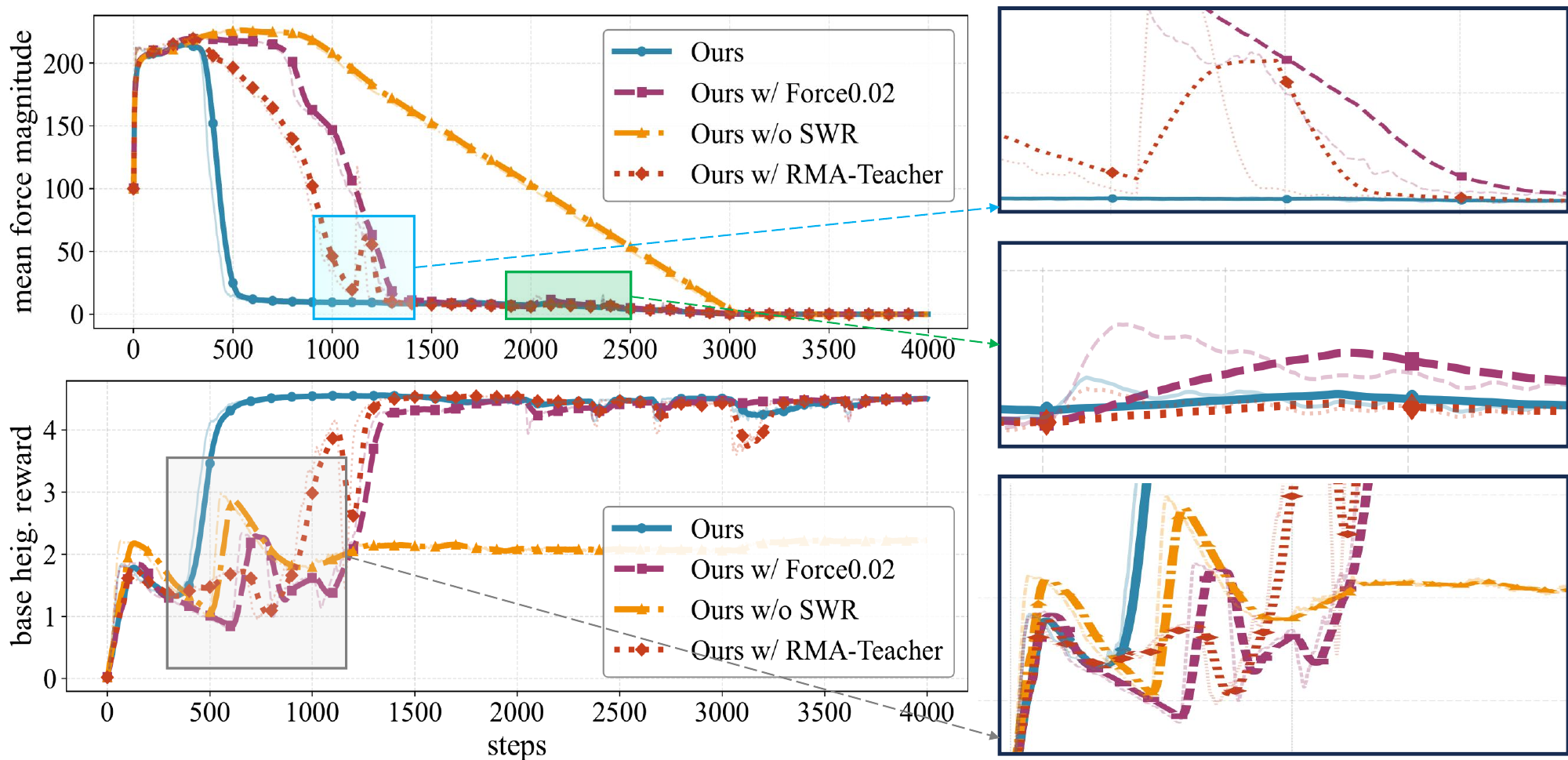}
\caption{\textbf{External force and height reward curves.} Smoother curves that reach their extremum (lowest/highest point) more rapidly indicate that the corresponding method produces a more stable policy and achieves faster training convergence.}
\label{fig: force-change-data}
\end{figure}

\begin{figure*}[t] 
  \centering
  \includegraphics[width=\textwidth]{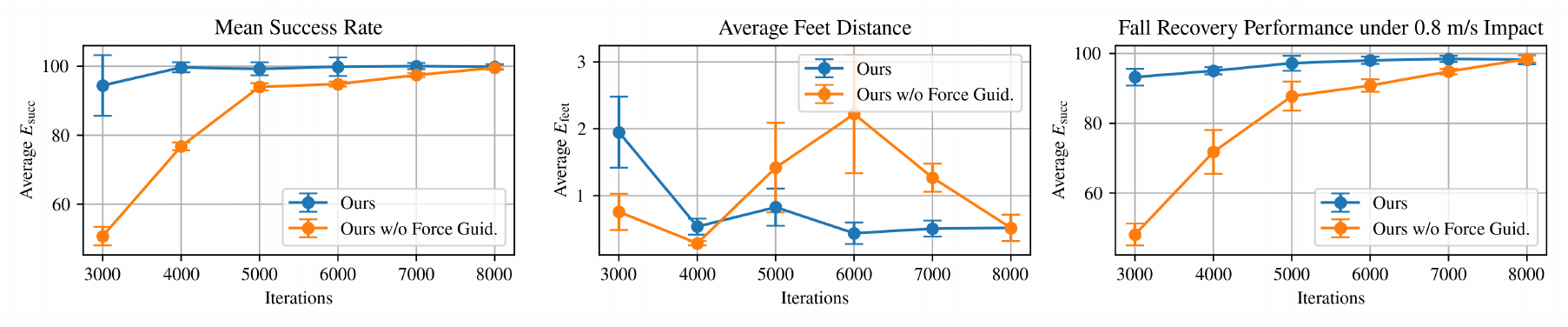} 
  \caption{\textbf{Success rate and foot travel distance across training iterations.} Compared with force‑free baselines, our force‑guided approach achieves a high recovery success rate by 3k iterations and attains improved stability by 4k iterations. These results demonstrate the rapid convergence and effectiveness of the proposed force-guided method.}
  \label{fig: iterations-succtest}
\end{figure*}

\subsection{Simulation Results}
Under identical reward functions and hyperparameter settings, we conducted comparative experiments across the following eight frameworks:

\begin{itemize}
\item \textbf{Ours}: Our full framework, which integrates force-guided with stage-wise rewards structure and a teacher-student training architecture.

\item \textbf{Ours w/ Force Curr.}: Here, the force-guided is replaced with a force curriculum learning strategy \cite{huang_learning_2025}, while keeping all other components unchanged.

\item \textbf{Ours w/o Force Guid.}: In this variant, the explicit force-guided term is removed from the training process. The robot is still subjected to an external force, but without the force-guided method.

\item \textbf{Ours w/ Force0.02.}: The baseline penalty multiplier is set to 0.001. To demonstrate that excessive Force-Guided leads to policy degradation, a comparative experiment is conducted with the multiplier increased to 0.02.

\item \textbf{Ours w/o SWR}: This variant removes the height-progressive stage-wise rewards design, utilizing only a single-layer reward signal $r^w$.

\item \textbf{Ours w/o T-S}: This variant abandons the teacher-student framework, reverting to a standard Actor-Critic structure.

\item \textbf{Ours w/ RMA Tea.}: This variant corresponds to the teacher model in \cite{kumar2021rma}, which embodies the optimal outcome within the teacher-student learning framework.

\item \textbf{HoST}: This refers to the method proposed in \cite{huang_learning_2025}.
\end{itemize}

Our framework enables armless bipedal-wheeled robots to recover from falls while maintaining sustained locomotion, and further generalizes to a humanoid platform. Ablation studies using metrics from \cite{huang_learning_2025} evaluate each component across varied terrains, with results summarized as follows:

\begin{figure}[!t]
\centering
\includegraphics[width=3.45in]{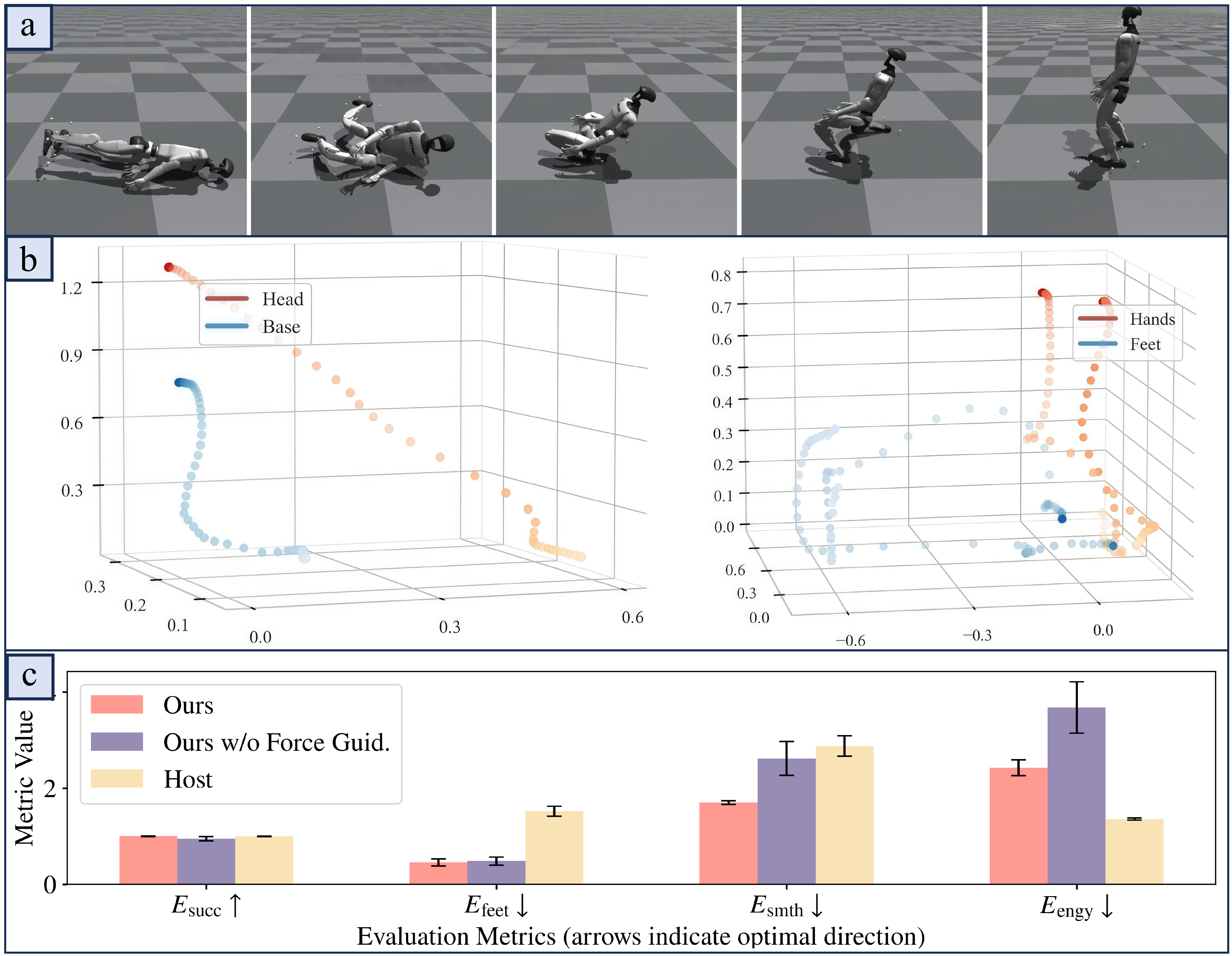}
\caption{\textbf{The proposed force‑guided method is extended to a 23‑DOF Unitree humanoid.} Our method learns a policy to stand up from arbitrary poses (a). The recovery motion is illustrated via 3D trajectories of key points (b), and a comparative evaluation (c) confirms the policy's smoothness and superiority over other methods.}
\label{tab: unitree test}
\end{figure}

\textbf{Height-progressive stage-wise rewards is crucial for guiding the recovery process.} Without the initial reward layers and adaptive target height updates, the force guided policy may rise with assistance but collapses upon force removal due to poor posture—evident in the pronounced force‑curve rebound during training (Fig. \ref{fig: force-change-data}, Table. \ref{tab:wheel_robot_sim_tab}), reflecting slow and unstable convergence. In contrast, the stage‑wise rewards policy attains stable locomotion at 0.8m/s and maintains a high recovery success rate while moving (Fig. \ref{fig: iterations-succtest}).

\textbf{Proper force-guided learning leads to more robust and efficient recovery strategies.} The conventional force-curriculum approach fails to achieve standing, while our redesigned force function enables rising but yields poor metrics (Table \ref{tab:wheel_robot_sim_tab}). In contrast, our method formulates force as optimizable constraints, driving rapid convergence to low-intervention and robust regions. Assistance is removed after 3k iterations, with stable, rapid force decay (Fig. \ref{fig: force-change-data}). The policy achieves force-free standing by 3k iterations, stabilizes after 4k with minimal foot adjustment, and maintains high success under commanded speeds (Fig. \ref{fig: iterations-succtest}).

\begin{figure*}[t] 
  \centering
  \includegraphics[width=\textwidth]{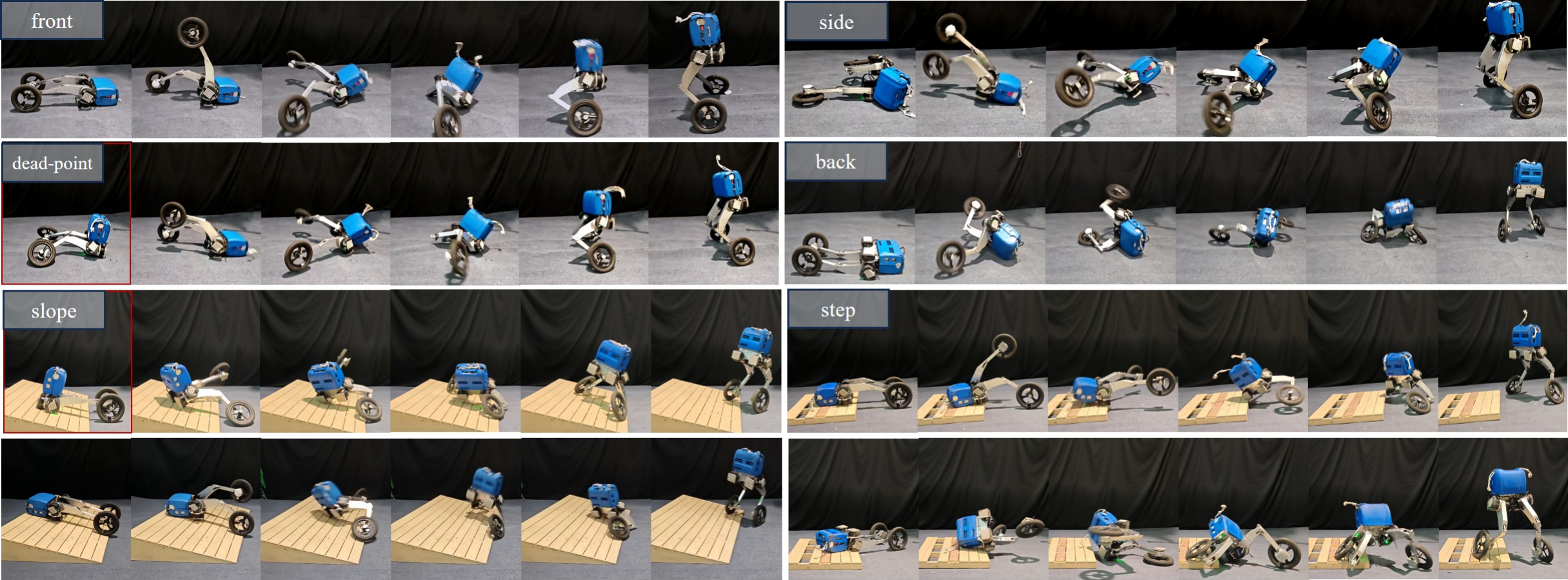} 
  \caption{\textbf{Standing-up tests under varied poses and terrains.}  The dead-point pose is highlighted with a box, which represents the final posture achieved by the force‑curriculum method. Our approach successfully overcomes this pose constraint and enables full standing recovery.}
  \label{fig: ground experiment}
\end{figure*}

\begin{figure*}[t] 
  \centering
  \includegraphics[width=\textwidth]{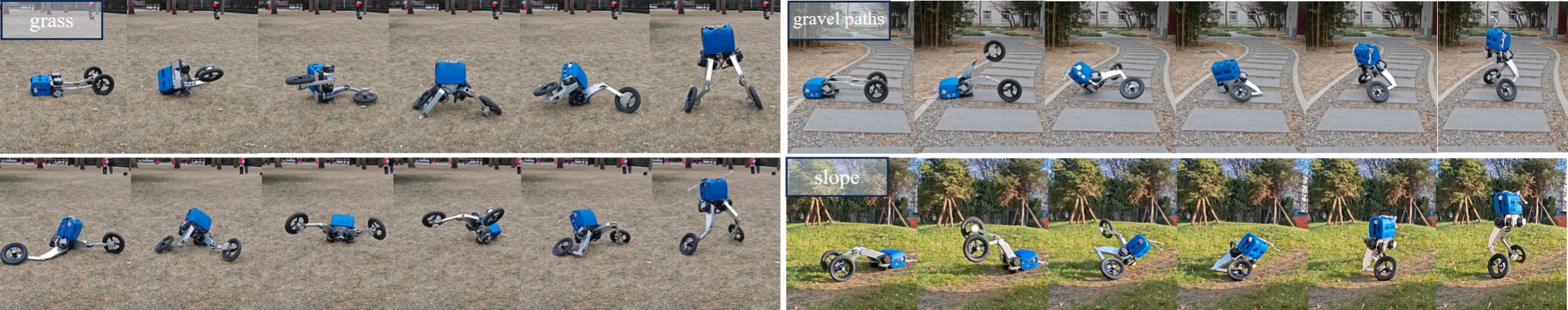} 
  \caption{\textbf{Standing-up tests outdoors under various initial poses.} These results demonstrate the robustness and environmental adaptability of our approach.}
  \label{fig: outdoor_total}
\end{figure*}

\textbf{Privileged information delivered through the teacher-student architecture improves recovery success.} Training with proprioception alone (\textbf{Ours w/o T‑S}) yields limited results, while incorporating privileged observations via the teacher-student framework significantly enhances performance, achieving comparable results to \textbf{Ours w/ RMA Tea.} (Table \ref{tab:wheel_robot_sim_tab}). This demonstrates that the teacher-student architecture effectively distills privileged knowledge of force effects and recovery dynamics, thereby improving the overall robustness and adaptability of the recovery process.
\par The method also generalizes to a 23‑DOF humanoid (Unitree), producing stable, smooth recovery and showing better foot control and motion smoothness than HoST after only 8k iterations (Fig. \ref{tab: unitree test}), further validating the FTSR framework.

\subsection{Real-World Experiments}

To validate the effectiveness of our approach, we deployed the trained policy on our self-developed $\textit{JiaRan}$ robot and conducted real-world tests across various terrains and initial poses. All experimental results demonstrate the robustness and adaptability of our method.

We evaluated the policy’s standing recovery across a range of initial poses (front, back, side, and dead‑point) and diverse terrains, including indoor flat ground, slope, and step (Fig. \ref{fig: ground experiment}), as well as outdoor grass, slope, and gravel paths (Fig. \ref{fig: outdoor_total}). The policy demonstrated strong environmental adaptability and robustness throughout these tests. Furthermore, quantitative success rates across outdoor terrains are reported in Table \ref{tab:success_rate_outdoor}, collectively confirming the method’s consistency, stability, and reliable adaptation in complex real‑world conditions.

\begin{figure*}[t] 
  \centering
  \includegraphics[width=\textwidth]{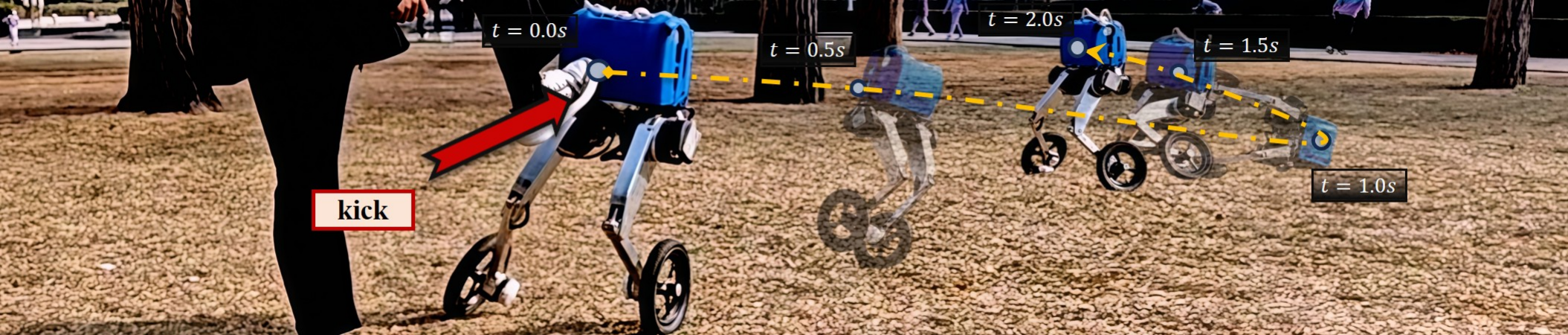} 
  \caption{\textbf{Fall recovery and subsequent locomotion test.} The direction of the external kick is marked with a red arrow, while the trajectory of the robot’s base during the experiment is indicated by a yellow dashed arrow. The results demonstrate that our method can autonomously recover from a fall and continue to execute motion commands.}
  \label{fig: fall-recovery total}
\end{figure*}

To demonstrate that the policy retains locomotor capability after recovery, we introduced kicking disturbances to the robot on grass and recorded the complete motion sequence, shown in Fig. \ref{fig: fall-recovery total}. It can be observed that after being pushed with substantial external force, the robot loses balance and falls, after which the policy autonomously drives it to recover a standing pose while preserving motion control ability thereafter.

\begin{table}[t]
\centering
\caption{Success Rates Across Different Outdoor Terrains}
\label{tab:success_rate_outdoor}
\begin{tabular}{l c c c c}
\toprule
\textbf{Metric} & \textbf{Grass} & \textbf{Slope} & \textbf{Stairs} & \textbf{Overall} \\
\midrule
Success count / Total trials & 5/5 & 5/5 & 5/5 & 15/15 \\
\bottomrule
\end{tabular}
\end{table}

\section{Conclusion}

This letter proposes the Force-guided Teacher-student framework with Stage-wise Rewards (FTSR). The framework enables bipedal-wheeled robots to recover from falls across multiple randomized and diverse terrains. By integrating a force-guided method, it addresses the trajectory-free fall recovery problem, ensuring stability and higher training efficiency. The teacher-student architecture combined with height-progressive stage-wise reward shaping utilizes privileged information through phased training, producing robust, high-success-rate policies. Comprehensive experiments validate the method’s robustness, adaptability, fast convergence, and stability on both simulated and physical bipedal-wheeled robots, also showing direct transfer potential to humanoids. Future work will test its utility on broader platforms.

\section*{Acknowledgment}
The authors would like to thanks to KinaMind Society for its technical support.



\bibliographystyle{IEEEtran}

\bibliography{ref}



\end{document}